\documentclass[sn-mathphys,Numbered]{sn-jnl}

\usepackage{graphicx}%
\usepackage{multirow}%
\usepackage{amsmath,amssymb,amsfonts}%
\usepackage{amsthm}%
\usepackage{mathrsfs}%
\usepackage[title]{appendix}%
\usepackage{xcolor}%
\usepackage{textcomp}%
\usepackage{manyfoot}%
\usepackage{booktabs}%
\usepackage{algorithm}%
\usepackage{algorithmicx}%
\usepackage{algpseudocode}%
\usepackage{listings}%

\usepackage[most]{tcolorbox}
\usepackage[disable]{todonotes}

\usepackage{color,soul}

\usepackage{fancyhdr}

\usepackage[caption=false]{subfig}

\usepackage{ulem}

\theoremstyle{thmstyleone}%
\theoremstyle{thmstyletwo}%

\theoremstyle{thmstylethree}%

\begin{document}

\title[Article Title]{Comparative Performance Evaluation of Large Language Models for Extracting Molecular Interactions and Pathway Knowledge}


\author*{\fnm{Gilchan} \sur{Park*}}\email{gpark@bnl.gov}

\author{\fnm{Byung-Jun} \sur{Yoon}}\email{\{byoon,xluo,vlopezmar,sjyoo,shantenu\}@bnl.gov}

\author{\fnm{Xihaier} \sur{Luo}}

\author{\fnm{Vanessa} \sur{López-Marrero}}
\author{\fnm{Shinjae} \sur{Yoo}}
\author{\fnm{Shantenu} \sur{Jha}}
 
\affil{\orgdiv{Computing and Data Sciences}, \orgname{Brookhaven National Laboratory}, \orgaddress{\street{\\PO Box 5000}, \city{Upton}, \postcode{11973}, \state{NY}, \country{USA}}}

\abstract{

\textbf{Background}\\ Identification of the interactions and regulatory relations between biomolecules play pivotal roles in understanding complex biological systems and the mechanisms underlying diverse biological functions. However, the collection of such molecular interactions has heavily relied on expert curation in the past, making it labor-intensive and time-consuming. To mitigate these challenges, we propose leveraging the capabilities of large language models (LLMs) to automate genome-scale extraction of this crucial knowledge.\\

\textbf{Results}\\ In this study, we investigate the efficacy of various LLMs in addressing biological tasks, such as the recognition of protein interactions, identification of genes linked to pathways affected by low-dose radiation, and the delineation of gene regulatory relationships. Overall, the larger models exhibited superior performance, indicating their potential for specific tasks that involve the extraction of complex interactions among genes and proteins. Although these models possessed detailed information for distinct gene and protein groups, they faced challenges in identifying groups with diverse functions and in recognizing highly correlated gene regulatory relationships.\\

\textbf{Conclusions}\\ By conducting a comprehensive assessment of the state-of-the-art models using well-established molecular interaction and pathway databases, our study reveals that LLMs can identify genes/proteins associated with pathways of interest and predict their interactions to a certain extent. Furthermore, these models can provide important insights, marking a noteworthy stride toward advancing our understanding of biological systems through AI-assisted knowledge discovery. 
\\\\
The code and data are available at: \href{https://github.com/boxorange/BioIE-LLM}{\texttt{https://github.com/boxorange/BioIE-LLM}}.
}




\maketitle

\section{Introduction}\label{sec1}

Understanding the intricate network of protein-protein interactions (PPIs), pathways, and gene regulatory relationships is crucial for deciphering cellular processes and disease mechanisms. In the pursuit of deeper insights into complex biological phenomena, an extensive array of heterogeneous data types has emerged from advanced experimental studies. The integration and analysis of such diverse data have garnered attention. Nevertheless, the interpretation of a voluminous and diverse dataset, coupled with the intrinsic noise in biological data, remains a significant challenge. An effective examination of omics data necessitates establishing causation and understanding the interplay of various factors, a task facilitated by the wealth of biological information embedded in scientific literature. This study endeavors to explore the potential efficacy of large language models (LLMs), with their vast parameter sizes and comprehensive training on extensive text corpora, hold great potential for automating information extraction related to biological tasks. The aim is to efficiently provide domain scientists with valuable information, requiring minimal human effort and time, thereby contributing to the enhancement of data interpretation in the field.

In the preliminary study \cite{park-etal-2023-automated}, we conducted an evaluation of the LLM named Galactica \cite{taylor2022galactica} with a focus on extracting protein interactions, pathway knowledge, and gene regulatory information. Building upon these initial findings, our present research extends these efforts by undertaking a comprehensive assessment and comparison of various LLMs. Notably, our work differs from the previous research in several key aspects:
\begin{enumerate}
    \item Model Evaluation Scope: We evaluated a total of 15 open-source LLMs, including recent state-of-the-art (SOTA) models, which are described in Section~\ref{sec3}.
    \item Negative PPI Samples: For the PPI task, we employed verified negative samples, enhancing the robustness of our evaluation (as presented in Section~\ref{sec3}).
    \item Pathway Selection Criteria: To recognize genes related to low-dose radiation (LDR), we meticulously selected human pathways affected by such exposure from the KEGG database (as described in Section~\ref{sec3}).
    \item In-Depth Result Analysis: Our study delved deeper into the results, utilizing various metrics to gain a comprehensive understanding of model performance.
    \item Improved Model Evaluation Framework: We enhanced the model evaluation framework by implementing distributed data parallelism, resulting in faster inference times.
\end{enumerate}
By rigorously examining LLMs in these contexts, we contribute to the advancement of our understanding of gene/protein functions and their relevance in life science research.

\section{Related Work}\label{sec2}

Diverse methodologies have been employed in the examination of PPIs, pathways, and relationships governing gene regulation.

Traditional statistical methods, while providing explicit inferences through defined probability models, often struggle with capturing the inherent complexity of biological systems. Techniques like Yeast Two-Hybrid (Y2H) \cite{bruckner2009yeast} offer insights into interactions within living cells, but suffer limitations like missing weak or transient interactions and generating false positives due to limited context. Bayesian networks, exemplified by tools like GeneNet \cite{ananko2002genenet} and bnlearn \cite{scutari2013bayesian}, leverage prior knowledge and data to infer relationships, but face challenges in computational demands, selection of accurate prior information, and handling noisy or biased data.

With the advent of high-throughput experimental technologies in genomics, transcriptomics, and proteomics, vast heterogeneous datasets have become readily available. This has necessitated the adoption of machine learning techniques, demonstrating superior performance in complex biological applications \cite{xu2019machine}. Notable applications include the analysis of protein structural properties \cite{vig2020bertology}, investigation of PPIs \cite{peng2017deep, park2022extracting}, and pathway analysis \cite{casani2020padhoc}. However, successful implementation often relies on fine-tuning models with large amounts of labeled data and domain-specific knowledge, requiring significant human effort and training time.

LLMs emerged as a compelling alternative for improving language processing efficiency in the biomedical domain \cite{Chen2023.04.19.537463}. LLMs have demonstrated promising advancements in addressing various Natural Language Processing (NLP) tasks within biomedicine, encompassing text generation, question answering (QA), and text summarization. Trained on extensive text corpora comprising web crawls, medical data, and  filtered and curated datasets, LLMs can process information from diverse sources such as scientific literature, databases, and various resources. This ability enables LLMs to capture nuanced relationships, context, and emerging knowledge often missed by traditional methods. Moreover, LLMs have the potential to prioritize candidate genes or pathways with minimal human intervention, saving researchers time and effort, especially when sifting through vast amounts of biomedical information.

Researchers investigated the potential of LLMs such as ChatGPT, Bard, and Claude for prioritizing and selecting genes based on existing knowledge, and the LLMs enabled them to efficiently analyze vast amounts of biomedical information, ultimately pinpointing candidate genes relevant to erythrocyte biology \cite{toufiq2023harnessing}. This study \cite{thapa2023chatgpt} explored the potential and limitations of LLMs like ChatGPT and Bard in the realm of biomedical research. The research findings highlighted that while LLMs may inadvertently produce misleading content, necessitating meticulous fact-checking and validation procedures, their capacity to efficiently analyze extensive scientific literature and suggesting novel research hypothesis connecting disparate concepts positions them as valuable tools.

In the present study, we evaluated the efficacy of 15 LLMs in extracting relevant information pertinent to biological tasks. Notably, the study prioritized open-source LLMs due to their enhanced flexibility and transparency compared to proprietary models, affording users greater customization and control over the models. The primary objective was to assess these models' effectiveness in retrieving insights from a corpus of biological literature and resources, thereby contributing to our understanding of LLMs' utility in biological research and informing potential applications in this domain.

\section{LLMs and Datasets}\label{sec3}

In this study, we explored the capabilities of open-source LLMs, namely Galactica, Alpaca, RST, Falcon, MPT, LLaMA2, Mistral, Mixtral, and SOLAR including smaller-sized biomedical domain-specific models like BioGPT and BioMedLM in tackling various biological tasks associated with PPIs, pathway knowledge, and gene regulatory relations. 

The reStructured Pre-training (RST) model \cite{yuan2022restructured} diverges from the prevailing decoder-only architecture commonly found in LLMs by adopting a Transformer encoder-decoder architecture. RST was trained on carefully designed data, which focuses on restructuring input and output data into specific formats to enhance pre-training efficacy. Meta AI developed an open source LLM, LLaMA \cite{touvron2023llama}, that has been trained on massive publicly available datasets. LLaMA models with much fewer parameters than strong competitors, such as GPT-3, Chinchilla, and PaLM, have outperformed these models on most benchmarks. However, one major drawback of LLaMA is that it is not well-suited for answering questions or following instructions. To address this limitation, a fine-tuned version of LLaMA, called Alpaca \cite{alpaca}, was trained on 52K instruction-following demonstrations. Alpaca behaves like conversational AI models, such as ChatGPT, and is able to answer questions and follow instructions. Falcon models \cite{falcon40b} have found extensive utility through training on a large-scale, high-quality English web corpus meticulously curated by Falcon Refined Web. MosaicML has introduced a publicly accessible and commercially viable series of MPT (MosaicML Pretrained Transformer) language model series \cite{team2023introducing}. The MPT-chat 70B model is noteworthy for its impressive ability to handle long-context inputs of up to 8K tokens. The Meta AI team has recently unveiled an enhanced iteration of the LLM, referred to as LLaMA2 \cite{touvron2023llama2}, which has undergone training with a significantly expanded dataset, amounting to 40\% more data compared to its predecessor, LLaMA, and featuring an extended context length. Mistral AI has introduced a compact-sized Mistral model featuring 7 billion parameters \cite{jiang2023mistral}, alongside its flagship model, Mixtral 8x7 model \cite{jiang2024mixtral}. Both models have integrated the Sliding Window Attention (SWA) mechanism, which adeptly handles longer sequences while preserving computational efficiency. Mixtral-8x7B, characterized by each layer comprising 8 feedforward blocks (experts), has employed a Sparse Mixture of Experts (SMoE) technique to expedite pretraining and enhance inference efficiency. Notably, Mixtral-8x7B accommodates a context of up to 32k tokens. In comparative evaluations, Mixtral demonstrates superior performance or parity with Llama 2 70B across various prominent benchmarks, all while utilizing significantly fewer active parameters during inference. The SOLAR-10.7B model \cite{kim2023solar}, developed by Upstage, represents an advanced LLM comprising 10.7 billion parameters. Employing a Transformer encoder-decoder structure, SOLAR-10.7B incorporates a novel methodology called depth up-scaling (DUS). This innovative approach combines architectural modifications and continued pretraining. Specifically, the model incorporates parameters from the Mistral 7B model into the upscaled layers, followed by continuous pretraining across the entire model. Despite its relatively compact design, this model demonstrates significant computational prowess, achieving SOTA performance that surpasses even larger models with parameter counts exceeding 30 billion. For the purposes of this study, we adopted the SOLAR-10.7B-Instruct model, which represents a fine-tuned version of SOLAR-10.7B tailored specifically for single-turn conversation tasks. Table~\ref{tab:llm-list} presents the technical specifications of the models.

\begin{sidewaystable}
\caption{A list of LLMs for the evaluation.}\label{tab:llm-list}
\begin{tabular*}{\textheight}{@{}lccccp{2.5in}}
\toprule%
Model & Release date & Developer & Parameters & Context length & Features \\
\midrule
\multirow{2}{*}{BioGPT-Large \cite{luo2022biogpt}} & \multirow{2}{*}{Feb-23} & \multirow{2}{*}{Microsoft} & \multirow{2}{*}{\underline{\textbf{1.5B}}} & \multirow{2}{*}{1024} & - Domain-specific foundation model \\
& & & & & - GPT-2 trained on biomedical literature for biological tasks \\
\midrule
\multirow{2}{*}{BioMedLM \cite{venigalla2022biomedlm}} & \multirow{2}{*}{Jan-23} & \multirow{2}{*}{Stanford} & \multirow{2}{*}{\textbf{2.7B}} & \multirow{2}{*}{1024} & - Domain-specific foundation model \\
& & & & & - GPT-2 trained on biomedical literature for medical question answering \\
\midrule
\multirow{2}{*}{Galactica \cite{taylor2022galactica}} & \multirow{2}{*}{Nov-22} & \multirow{2}{*}{Meta} & 120M, 1.3B, \underline{\textbf{6.7B}}, & \multirow{2}{*}{2048} & - Trained on scientific literature \\
& & & \underline{\textbf{30B}}, 120B & & - Designed data for scientific tasks \\
\midrule     
Alpaca \cite{alpaca} & Mar-23 & Stanford & \underline{\textbf{7B}} & 2048 & - Instruction fine-tuned version of the LLaMA 7B model on 52K instruction-following demonstrations \\
\midrule
\multirow{2}{*}{RST \cite{yuan2022restructured}} & \multirow{2}{*}{Sep-22} & \multirow{2}{*}{CMU} & \multirow{2}{*}{\underline{\textbf{11B}}} & input: 1024 & - Transformer encoder-decoder framework \\
& & & & output: 256 & - Designed data for various NLP tasks \\
\midrule
Falcon \cite{falcon40b} & Mar-23 & TII & \underline{\textbf{7B, 40B}} & 2048 & - Trained on high-quality data filtered by Falcon RefinedWeb \\
\midrule
\multirow{2}{*}{MPT-Chat \cite{team2023introducing}} & \multirow{2}{*}{Jul-23} & \multirow{2}{*}{MosaicML} & \underline{\textbf{7B}} & 2048 & \multirow{2}{*}{- A chatbot-like MPT model for dialogue generation} \\
& & & \underline{\textbf{30B}} & 8192 & \\
\midrule
\multirow{2}{*}{Llama-2-Chat \cite{touvron2023llama2}} & \multirow{2}{*}{Jul-23} & \multirow{2}{*}{Meta} & \multirow{2}{*}{\underline{\textbf{7B}}, 13B, \underline{\textbf{70B}}} & \multirow{2}{*}{4096} & - 40\% more data than Llama 1 and has double the context length \\
& & & & & - S fine-tuned version of Llama 2 that is optimized for dialogue use cases \\
\midrule
\multirow{2}{*}{Mistral-Instruct \cite{jiang2023mistral}} & \multirow{2}{*}{Sep-23} & \multirow{2}{*}{Mistral AI} & \multirow{2}{*}{\underline{\textbf{7.3B}}} & \multirow{2}{*}{8192} & - Sliding Window Attention (SWA) mechanism \\
& & & & & - Instruction fine-tuned version of the Mistral 7B model \\
\midrule
\multirow{2}{*}{Mixtral-8x7-Instruct \cite{jiang2024mixtral}} & \multirow{2}{*}{Dec-23} & \multirow{2}{*}{Mistral AI} & \multirow{2}{*}{\underline{\textbf{46B}}} & \multirow{2}{*}{32768} & - SMoE (Sparsed Mixture of Experts) \\
& & & & & - Instruction fine-tuned version of the Mistral-8x7B model \\
\midrule
\multirow{4}{*}{SOLAR-Instruct \cite{kim2023solar}} & \multirow{4}{*}{Dec-23} & \multirow{4}{*}{Upstage} & \multirow{4}{*}{\underline{\textbf{10.7B}}} & \multirow{4}{*}{4096} & - Transformer encoder-decoder architecture \\
& & & & & - Depth up-scaling (DUS) \\
& & & & & - Leveraged weights from the Mistral 7B model \\
& & & & & - SOLAR-10.7B Fine-tuned version for single-turn conversation \\
\botrule
\end{tabular*}
\footnotetext{Note: \underline{\textbf{Bold and underlined models}} were evaluated in this study. The largest model, Galactica 120B, could not be tested due to computational power constraints.}
\end{sidewaystable}

To create the biological tasks, STRING \citep{szklarczyk2021string}, Negatome \citep{blohm2014negatome}, KEGG \citep{kanehisa2000kegg}, INDRA \citep{bachman2023automated} databases were adopted. Detailed descriptions of the STRING, KEGG, and INDRA databases are provided in our previous paper \citep{park-etal-2023-automated}, and the description of the Negatome database can be found in the Appendix~\ref{secA1:db-desc}. In the investigation of negative PPI pairs, the initial work \citep{park-etal-2023-automated} employed unlinked protein pairs from the STRING database to represent non-interacting proteins. However, subsequent analysis raised concerns about potential false negatives within this dataset, arising from undetected interactions absent from the latest database updates. To ensure reliable negative interactions, we opted for experimentally validated non-interacting protein pairs from Negatome 2.0. This study also utilized newly and meticulously selected KEGG pathways affected by Low-Dose Radiation (LDR) exposure. KEGG consolidates genomic data in the GENES database, encompassing gene catalogs from fully and partially sequenced genomes, annotated with current gene functions. The pathway database enriches this genomic information by integrating higher-order functional data with ortholog group tables. These tables facilitate the identification of conserved subpathways, encoded by genes that are often positionally related on the chromosome, offering invaluable insights into gene function prediction. Specifically, from the 548 available KEGG pathway maps, we identified 343 pathways pertinent to our gene expression dataset, GSE43151. Pathways lacking any genes measured in GSE43151 were excluded from our analysis.

The next step is to identify meaningful pathways for analysis. To achieve this, we leveraged the gene expression dataset GSE43151 to examine pathway activities relevant to different radiation exposures: zero-dose, low-dose, and high-dose. Assuming Gaussian distributions for gene expression levels, we calculated the log-likelihood ratio (LLR) for each gene within a pathway to discern expression patterns indicative of the specific radiation exposure level. The aggregate LLR across genes in a pathway provided a measure of the pathway's activity level, indicative of the phenotype. Given the potential variability in expression data, we normalized the LLR values to enhance the robustness of our analysis. This normalization diverges from traditional Naïve Bayes models, offering a refined approach to infer pathway activity. To assess the discriminative power of pathways between different radiation exposures, we employed $t$-test statistics on the normalized activity levels and computed an aggregated differential activity score for each pathway. Our methodology culminated in the ranking of KEGG pathways based on their differential activity scores, comparing zero-dose against low-dose, and zero-dose against high-dose samples. This approach allowed us to identify and characterize the most significantly impacted biological pathways under varying radiation levels.

\section{Experiment}\label{sec4}
We conducted a comprehensive evaluation by comparing the LLMs in question answering formatted tasks. In the context of LLMs, the proper selection of the number of examples or shots is essential to ensure efficient engineering. For this purpose, various number of shots ranging from zero to five were examined to determine the most effective quantity of shots for each specific task, and the shot number yielding the best performance outcome was documented in the results. Additionally, prompt construction is another critical factor that merits attention, and the prompts tested for each task are listed in Appendix~\ref{secA2:tested-prompt}. The experiments were conducted on 4$\times$NVIDIA A100 80GB GPUs. The model processed a batch sized input for a task, which is the number of prompts to infer (I.e., the number of input texts for model generation at once). For this study, we established a testing infrastructure utilizing the HuggingFace framework. To enhance time efficiency, we employed data parallel inference techniques, facilitating the concurrent processing of batched inputs across multiple GPUs. Detailed information regarding the task execution duration is presented in the Appendix~\ref{secA3:task-exec-time}.

\subsection{Recognizing Protein-Protein Interactions (PPIs)}
We assessed the performance of the LLMs in identifying protein binding information using a human protein network obtained from the STRING DB. Our main focus was on using the models to generate a list of proteins that interact with a given protein, as part of the generative question task (\textit{\textbf{STRING DB PPI Task: generative question}}). Box 1 illustrates an example of this task. In this box, the upper section contains a list of actual proteins, while the lower section presents a list of proteins predicted by a LLM. Text highlighted in \textcolor{blue}{blue} indicates matching information, whereas text highlighted in \textcolor{red}{red} denotes discrepancies.

\begin{tcolorbox}[title=Box 1: Example of STRING DB PPI Task: generative question, label=box1]
Question: Which proteins interact with ARAP1?\\
Answer (true): \textcolor{blue}{CDC42, ARAP3, CLIP4, ARAP2, CLASP2,}
\textcolor{red}{IQGAP1, RAB6A, DOK2, CLTA, KCNQ1}
\tcblower
Answer (pred): \textcolor{blue}{CDC42, ARAP3, CLIP4, ARAP2, CLASP2,}
\textcolor{red}{CLASP1, DCTN8, DCTN14, CEP290, CSPP1}
\end{tcolorbox}

In order to assess performance, we randomly selected 1,000 proteins and, for each protein, compared 10 generated proteins by a model with true binding proteins, resulting in the evaluation of a total of 10,000 PPI pairs. The reason for measuring only 10 binding proteins is that each protein sourced from the STRING database often has an extensive list of interacting proteins. While we examine the complete interacting protein list for each sample protein, the constraints imposed by the model's maximum length for generation and the need for efficient inference necessitated the evaluation of only 10 interacting proteins for each protein. For instance, if a protein A exhibits interactions with 100 other proteins, we examined whether the 10 proteins predicted by the model are present within this set of 100 interacting proteins. Our evaluation criteria included micro F1, macro F1, and the count of fully matched proteins out of the initial 1,000 protein candidates. Micro F1 was employed to gauge matches across all 10,000 pairs, while macro F1 quantified matches for each protein label used as a query like the exemplar of ARAP1 provided earlier. The count of fully matched proteins served as an indicator of the depth of the models' knowledge concerning specific proteins, with an illustration of a fully matched protein presented in Box 2.

\begin{tcolorbox}[title=Box 2: A fully matched example of STRING DB PPI Task: generative question]
Question: Which proteins interact with EED?\\
Answer (true): \textcolor{blue}{HDAC1, SMARCA4, HMGB2, CBX5, HDAC2, EZH2, CBX3, GATA2, STAG2, RB1}
\tcblower
Answer (pred): \textcolor{blue}{HDAC1, SMARCA4, HMGB2, CBX5, HDAC2, EZH2, CBX3, GATA2, STAG2, RB1}
\end{tcolorbox}

\begin{table}[h]
\caption{STRING DB PPI Task - Model performance for the generated binding proteins for 10K protein pairs from 1K protein list.}\label{tab:ppi-task1-eval}%
\begin{tabular}{@{}lccc@{}}
\toprule
 & Micro F1 & Macro F1 & \# Full Match out of 1K protein list \\
\midrule
BioGPT-Large (1.5B) & 0.1220 & 0.1699 & 10 \\
BioMedLM (2.7B)     & 0.1598 & 0.1992 & 61 \\
Galactica (6.7B)	& 0.2110 & 0.2648 & 75 \\
Galactica (30B)	    & 0.2867 & 0.3516 & 110 \\
Alpaca (7B)	        & 0.0998 & 0.1388 & 16 \\
RST (11B)	        & 0.0987 & 0.1523 & 10 \\
Falcon (7B)	        & 0.0435 & 0.0632 & 7 \\
Falcon (40B)        & 0.1246 & 0.1607 & 35 \\
MPT-Chat (7B)	    & 0.1313 & 0.1658 & 45 \\
MPT-Chat (30B)	    & 0.2926 & 0.3467 & 144 \\
LLaMA2-Chat (7B)	& 0.2807 & 0.3498 & 89 \\
LLaMA2-Chat (70B)	& 0.3517 & 0.4187 & 159 \\
Mistral-Instruct (7B)    & 0.2762 & 0.3299 & 126 \\
Mixtral-8x7B-Instruct (46B)	& \textbf{0.3867} & \textbf{0.4295} & \textbf{258} \\
SOLAR-Instruct (10.7B)  & 0.2766 & 0.3260 & 141 \\
\botrule
\end{tabular}
\footnotetext{Note: 5-shot prompting was used for the evaluation. \textbf{Bold} indicates the best score.}
\end{table}

The model performance results are presented in Table~\ref{tab:ppi-task1-eval}. Among the models, Mixtral-8x7B-Instruct (46B) demonstrated the most accurate predictions followed by LLaMA2-Chat (70B), and Mistral-Instruct (7B) and LLaMA2-Chat (7B) exhibited performance levels comparable to larger models, such as MPT-Chat (30B) and Galactica (30B). In contrast, the Falcon and RST models displayed comparatively poorer performance. Particularly it is noteworthy that Falcon (40B) significantly underperformed relative to other larger models. The discrepancy between macro F1 scores surpassing micro F1 scores suggests a nuanced understanding of particular proteins within the models, contrasting with their broader comprehension of overall protein characteristics. This observation implies that larger models may harbor an augmented repository of intricate insights concerning specific proteins, as evidenced by the prevalence of proteins exhibiting a full complement of 10 binding protein matches.

For the analysis of model predictions, we identified a set of proteins that were consistently predicted by the top five scoring models across all 10 pairs given that no proteins that all models predicted 10 related pairs exist. This set includes WNT7B, CCND1, EIF3L, and ITGAM. Conversely, there exists another set of proteins that were not predicted by any of the models, including STKLD1, TMEM91, CXorf38, SFSWAP, SMIM34A, TMEM89, TMEM123, ZNF674, FAM218A, GMNC, LUZP6, ZMYND19, ENSG00000275217, EAPP, ZNF396, PSTK, ZNF641, PRR22, ABRACL, ENSG00000267561, KIAA1751, VIT, MDFIC, TRAM1L1, C11orf94, RABL3, RSBN1L, TMEM189, ZNF581, ENSG00000263020, FAM159A, ZNF385A, MRGPRX4, RGP1, FAM180A, C2orf68, RBM18, GRAMD2, ZSCAN25, KIAA0895, DCAF4L2, CXorf40B, CXorf66, ENSG00000198590, C5orf47, ZNF835, MANSC4, C15orf61, FSD1L, YRDC, URM1, ZNF787, RMND1, RBM44, PRR34, C5orf58, MAGEF1, AXDND1, SMIM1, TMEM74, TMEM217, ENSG00000267881, SPATA4, SMIM2, ZNF829, ENSG00000271786, ZNF474, TMEM88, ZNF839, TEX30, CCDC43, C5orf15, OCLM, ENSG00000183628, C1orf52, C19orf70, SDR39U1, LEPROTL1.

It is noteworthy that these two sets of proteins belong to distinct functional categories and exhibit differential roles in cellular processes. The proteins within the former set are better characterized and play roles in cell signaling, cell cycle regulation, and immune responses. Notably, these proteins are associated with regulatory functions related to diseases. For instance, Mutations in EIF3L are associated with conditions such as diaphyseal medullary stenosis with malignant fibrous histiocytoma, High WNT7B expression is associated with poor prognosis in cancer patients, ITGAM is implicated in various immune processes and is associated with diseases such as Systemic Lupus Erythematosus (SLE), and CCND1 is considered an oncogene, playing critical roles in cell proliferation, growth, angiogenesis, and resistance to chemotherapy and radiotherapy. On the other hand, the proteins within the latter set exhibit diverse functions, including enzyme activity, signal transduction, or transcriptional regulation, albeit their specific associations with diseases remain less well-established. These proteins have varied functions and require further investigation to fully understand their roles in cellular processes. The observed disparity in the predictive capacity of the models can be attributed to the composition of the two protein sets. The first set predominantly comprises proteins that have undergone rigorous investigation and possess well-defined roles in diverse biological processes. These extensively studied proteins are documented in scientific databases and publications, making relevant information readily accessible. In contrast, the second set primarily consists of uncharacterized proteins for which limited or no data regarding their structures, functions, or interactions are available. Consequently, the lack of comprehensive knowledge about this latter group contributes to the models' reduced predictive performance in their case.

Following that, we conducted an evaluation to assess the model's ability to recognize protein binding relationships in a binary framework. Specifically, we formulated a yes/no inquiry aimed at determining the existence of any association or interaction between two proteins (\textit{\textbf{STRING/Negatome DB PPI Task: yes/no question}}). Boxes 3 and 4 depict examples of this task.

\begin{tcolorbox}[title=Box 3: Example of STRING/Negatome DB PPI Task: yes/no question]
Question: Do TMEM43 and POTEI interact with each other?\\
Answer (true): \textcolor{blue}{yes}
\tcblower
Answer (pred): \textcolor{blue}{yes}
\end{tcolorbox}

\begin{tcolorbox}[title=Box 4: Example of STRING/Negatome DB PPI Task: yes/no question]
Question: Do Q5JTD0 and A5JSJ9 interact with each other?\\
Answer (true): \textcolor{blue}{no}
\tcblower
Answer (pred): \textcolor{blue}{no}
\end{tcolorbox}

For the experiment, we randomly selected 1,000 protein pairs from each STRING database and Negatome database, resulting in the evaluation of a total of 2,000 samples (1,000 positive PPIs and 1,000 negative PPIs). The performance of the models is detailed in Table~\ref{tab:ppi-task2-eval}, and the corresponding confusion matrix is illustrated in Appendix~\ref{secA4:ppi-confusion-matrix}. Notably, the chat-based LLMs showed superior performance in this yes/no question answering task. Specifically, MPT-Chat (7B) demonstrated the most favorable performance followed by LLaMA-2-Chat (70B). On the contrary, Galactica (6.7B) and BioGPT-Large exhibited diminished performance levels, with Falcon models manifested almost zero capability in responding to questions. A potential explanation for the observed performance gap in binary question answering tasks may lie in two factors: 1) the inherent lack of domain-specific information within the training data, and 2) the potential inadequacy of model parameters to capture and exploit such information effectively. Consequently, this limitation may impede the model's ability to comprehend and address the domain-specific binary question answering formats effectively, thereby hindering its capacity to extract the desired responses. Specifically, Falcon models primarily undergo training using extensive textual corpora, including web-based documents and literary works. However, these sources may not emphasize the acquisition of factual knowledge essential for proficient binary question-answering tasks. Additionally, Falcon models might not have encountered diverse and comprehensive question-answering datasets, which could contribute to suboptimal performance.

\begin{table}[h]
\caption{STRING/Negatome DB PPI Task - F1 scores for randomly selected 2,000 (1,000 positive + 1,000 negative) PPI pairs.}\label{tab:ppi-task2-eval}%
\begin{tabular}{@{}p{7cm}cc@{}}
\toprule
 & Micro F1 (\#shot) & Macro F1 (\#shot) \\
\midrule
BioGPT-Large (1.5B) & 0.5700 (1-shot) & 0.4811 (1-shot) \\
BioMedLM (2.7B)     & 0.7125 (2-shot) & 0.6866 (2-shot) \\
Galactica (6.7B)	& 0.5320 (1-shot) & 0.4568 (1-shot) \\
Galactica (30B)	    & 0.8585 (5-shot) & 0.8585 (5-shot) \\
Alpaca (7B)	        & 0.6660 (5-shot) & 0.6241 (5-shot) \\
RST (11B)	        & 0.6990 (0-shot) & 0.6701 (0-shot) \\
Falcon (7B)	        & 0.5000 (1-shot) & 0.3333 (1-shot) \\
Falcon (40B)        & 0.5050 (1-shot) & 0.3443 (1-shot) \\
MPT-Chat (7B)	    & \textbf{0.9795} (5-shot) & \textbf{0.9795} (5-shot) \\
MPT-Chat (30B)	    & 0.9345 (5-shot) & 0.9343 (5-shot) \\
LLaMA2-Chat (7B)	& 0.8670 (5-shot) & 0.8662 (5-shot) \\
LLaMA2-Chat (70B)	& 0.9545 (5-shot) & 0.9545 (5-shot) \\
Mistral-Instruct (7B)    & 0.7745 (5-shot) & 0.7707 (5-shot) \\
Mixtral-8x7B-Instruct (46B)	& 0.7770 (5-shot) & 0.7658 (5-shot) \\
SOLAR-Instruct (10.7B)  & 0.7615 (3-shot) & 0.7481 (3-shot) \\
\botrule
\end{tabular}
\footnotetext{Note: \textbf{Bold} indicates the best score.}
\end{table}

We conducted an analysis of predictions generated by all models, excluding Galactica (6.7B), BioGPT-Large, and Falcon series due to their propensity to produce biased results. All questions were addressed by the models, with each question receiving a correct prediction from at least one model. Positive PPIs consistently predicted by all models include (ACOT2, HADHA), (ABCB11, HSPA5), and (ADAM21, MMP24), reflecting functionalities spanning fatty acid metabolism, bile acid transport, endoplasmic reticulum stress response, and extracellular matrix remodeling. Conversely, negative PPIs consistently predicted by all models included various pairs: (P17036, Q8TBX8), (P49459, Q9NW38), (P04141, Q6NSJ8), (O15350, Q76N89), (Q02535, Q70SY1), (P16104, Q8N423), (P41240, Q9UH92), (P36915, Q9UH92), (P25963, Q02535), (P27361, Q49A26), (P84022, Q13547), (P17036, Q9UH92), (P36915, Q8TBX8), (Q07157, Q5JTD0), (P48729, Q96EV8), (Q13077, Q8TDR0), (P25963, Q8TBX8), (P42224, Q9Y6X2), (O43353, Q5XLA6), (P13861, Q9ULX6), (Q13239, Q6PIZ9), (P18847, Q8TBX8), (Q13642, Q5TD97), (P26038, Q9H204), (P06400, Q9NX02), (Q06265, Q5RKV6), (P83436, Q14746), (P17036, Q9UHR5), (P18848, Q8TBX8), (P22681, Q6PIZ9), (Q05519, Q8WU68), (P17036, Q68DY1), (Q12962, Q6P1X5), (P24941, Q9UBD5), (Q16526, Q8NEZ5), (P17036, Q70SY1), (P36915, Q68DY1), (P31150, Q70SY1), (P15976, Q05513), (P15514, Q969F2), (P13765, Q6ICR9), (Q13309, Q49AN0), (Q15406, Q9Y618), (Q13571, Q6PIZ9). Notably, certain proteins, like P17036 and Q8TBX8, recurred across multiple negative interactions. While the identified negative pairs exhibited interactions for diverse reasons and under various biological conditions, some pairs shared similar functionalities. For instance, Q05519 and Q8WU68 are both involved in the regulation of transcription, Q12962 and Q6P1X5 play roles in cell proliferation regulation, and P13765 and Q6ICR9 are associated with apoptosis regulation.

\subsection{Identifying Genes related to Human Pathways affected by Low-dose Radiation (LDR) Exposure}
This experiment aimed to assess the models' capacity to identify genes associated with human pathways relevant to LDR exposure in the KEGG database. The objective of the task was to generate a comprehensive list of genes that are part of human pathways specifically connected to LDR exposure (\textit{\textbf{KEGG DB Pathways affected by LDR exposure Task: generative question}}). Boxes 5 and 6 show examples of this task.

\begin{tcolorbox}[title=Box 5: Example of KEGG DB Pathways affected by LDR exposure Task: generative question]
Question: Which genes are associated with ``Porphyrin and chlorophyll \\metabolism"?\\
Answer (true): \textcolor{blue}{HMBS, HMOX1, CPOX, FECH, PBGD, ALAS1, ALAS2,} \textcolor{red}{HMOX2, GLUPRORS, UGT1}
\tcblower
Answer (pred): \textcolor{blue}{HMBS, HMOX1, CPOX, FECH, PBGD, ALAS1, ALAS2,} \textcolor{red}{MCOPS7, MLS, LSDMCA1}
\end{tcolorbox}

\begin{tcolorbox}[title=Box 6: Example of KEGG DB Pathways affected by LDR exposure Task: generative question]
Question: Which genes are associated with ``Nicotine addiction"?\\
Answer (true): \textcolor{blue}{NR2A, NR2B, GABRA2, GRIN2A, VGAT,} \textcolor{red}{GABRP, GluN1, GLURC, EIEE43, NMDAR2B}
\tcblower
Answer (pred): \textcolor{blue}{NR2A, NR2B, GABRA2, GRIN2A, VGAT,} \textcolor{red}{GRM8, COMT, DAO, CYP3A4, HPA}
\end{tcolorbox}

In our experiments, we chose the top 100 pathways exhibiting the most significant differential activation in response to LDR exposure. For each pathway, we compared 10 genes predicted by a model with the actual genes associated with the pathway. The prediction performance of the models on the genes associated with the pathways is presented in Table~\ref{tab:pathway-task-eval}. Mixtral-8x7B (46B) most accurately predicted the genes related to the pathways followed by BioMedLM, Galactica (30B), MPT-Chat (30B), and SOLAR-Instruct (10.7B) models, whereas the Alpaca and RST models showed the worst performances. Notably, the overall performance of the models surpassed that of the previous generative test conducted for \textit{\textbf{STRING DB PPI Task}}. One possible explanation for the models' enhanced ability to recognize pathways linked to LDR exposure, compared to proteins, is that pathway names specifically associated with LDR are often mentioned in narrower and specific sections or categories within the literature. In contrast, protein names are more commonly dispersed across a wider range of topics in scientific papers. That suggests that models' search for information within a clearly delineated collection of data may yield more precise outcomes with less hallucinations compared to searching for information derived from ambiguous inputs sourced from heterogeneous sources. This might also account for the reason that BioMedLM and BioGPT-Large exhibited the significant improvement in generating correct list compared to \textit{\textbf{STRING DB PPI Task}}, and the domain-specific models outperformed some of the larger language models trained on more diverse datasets. As delineated in the prior study \cite{park-etal-2023-automated}, the models demonstrate a propensity to produce predictions closely resembling actual names. Some examples are illustrated in Appendix~\ref{secA5:kegg-task-pred-ex}.

\begin{table}
\caption{KEGG DB Pathways affected by LDR exposure Task - Model performance for 998 genes that belong to the top 100 pathways associated with low-dose radiation exposure.}\label{tab:pathway-task-eval}
\begin{tabular}{@{}lccc@{}}
\toprule
 & Micro F1 (\#shot) & Macro F1 (\#shot) & \# Full Match out of 100 \\
\midrule
BioGPT-Large (1.5B) & 0.2435 (3-shot) & 0.3131 (3-shot) & 5 \\
BioMedLM (2.7B)     & 0.4619 (2-shot) & 0.5383 (2-shot) & 22 \\
Galactica (6.7B)	& 0.3136 (5-shot) & 0.3874 (5-shot) & 8 \\
Galactica (30B)	    & 0.4609 (5-shot) & 0.5304 (5-shot) & 24 \\
Alpaca (7B)	        & 0.1172 (3-shot) & 0.1439 (3-shot) & 4 \\
RST (11B)	        & 0.1102 (3-shot) & 0.1238 (3-shot) & 7 \\
Falcon (7B)	        & 0.1393 (3-shot) & 0.1681 (3-shot) & 5 \\
Falcon (40B)        & 0.2004 (3-shot) & 0.2367 (3-shot) & 7 \\
MPT-Chat (7B)	    & 0.1894 (5-shot) & 0.2482 (5-shot) & 4 \\
MPT-Chat (30B)	    & 0.3978 (5-shot) & 0.4550 (5-shot) & 18 \\
LLaMA2-Chat (7B)	& 0.2936 (5-shot) & 0.3874 (5-shot) & 8 \\
LLaMA2-Chat (70B)	& 0.3908 (5-shot) & 0.4577 (5-shot) & 18 \\
Mistral-Instruct (7B)	& 0.3828 (2-shot) & 0.4416 (2-shot) & 19 \\
Mixtral-8x7B-Instruct (46B)	& \textbf{0.5962} (2-shot) & \textbf{0.6479} (2-shot) & \textbf{39} \\
SOLAR-Instruct (10.7B)	& 0.3928 (2-shot) & 0.4537 (2-shot) & 19 \\
\botrule
\end{tabular}
\footnotetext{Note: \textbf{Bold} indicates the best score.}
\end{table}

In our investigation of model predictions, we initially assessed the entirety of pathways accurately predicted by all models. Given that no pathway was entirely predicted by every model, our focus shifted to towards identifying common pathways among the top five performing models, each of which successfully predicted all ten pairs. The pathways encompassed by these models include ``GABAergic synapse", ``Metabolism of xenobiotics by cytochrome P450', and ``Glycerolipid metabolism'. Conversely, there were pathways that remained unpredicted by all models. These pathways include ``Neomycin", ``kanamycin and gentamicin biosynthesis", ``Selenocompound metabolism", and ``Riboflavin metabolism". The fully predicted pathway group primarily focuses on cellular communication (GABAergic synapse), detoxification (cytochrome P450), and lipid metabolism (glycerolipids), and dysregulation in these pathways may impact neurological function, drug metabolism, and lipid-related disorders. The non-predicted pathways relates to antibiotic production (neomycin, kanamycin, and gentamicin), selenium metabolism, and vitamin B2 utilization, and these pathways are relevant to antibiotic resistance, selenium deficiency, and riboflavin-related health conditions. The differential predictive capacity of the models can be elucidated by more extensively studied research in the fully predicted pathway group, owing to its substantial implications for health and disease. In contrast, the non-predicted pathway group has received less attention overall. While they are essential, their research scope tends to be narrower compared to the broader implications of the first group. Upon scrutinizing the predictions of each model, it became evident that each model encompasses varying depths of information concerning specific pathways, as discerned from the distinct count of uniquely identified fully matched pathways. The list of unique pathways that received full prediction coverage by models can be found in Appendix~\ref{secA6:kegg-task-unique-full-list}.

\subsection{Evaluating Gene Regulatory Relations}
In this evaluation, we assessed the models’ proficiency in discerning human gene regulatory relationships. To achieve this, we employed data sourced from the INDRA database~\cite{bachman2023automated}. The INDRA data comprises text statements extracted from scientific research papers, thereby providing contextual information about the entities involved in these relationships. Leveraging these text snippets, we formulated questions for the models. Specifically, we tasked the models with selecting the accurate relationship between two genes from various relation classes within a given text. This evaluation employs a multiple-choice question format to assess the models' ability to predict gene regulatory relationships using gene information as well as their proficiency in reading comprehension, specifically within the domain of gene regulatory relation texts (\textit{\textbf{INDRA DB Gene Regulatory Relation Task: Multiple-choice question}}). Examples of this task are presented in Boxes 7 and 8.

\begin{tcolorbox}[title=Box 7: Example of INDRA DB Gene Regulatory Relation Task: Multiple-choice question]
Context: In 2006, we demonstrated that activation of TRPM2 appeared to induce insulin secretion.\\\\
Question: Given the options: ``Activation", ``Inhibition", ``Phosphorylation", ``Dephosphorylation", ``Ubiquitination", ``Deubiquitination", which one is the relation type between TRPM2 and insulin in the text above?\\\\
Answer (true): \textcolor{blue}{Activation}
\tcblower
Answer (pred): \textcolor{blue}{Activation}
\end{tcolorbox}

\begin{tcolorbox}[title=Box 8: Example of INDRA DB Gene Regulatory Relation Task: Multiple-choice question]
Context: WRN was shown to genetically interact with topoisomerase 3 and restore the slow growth phenotype of sgs1 top3.\\\\
Question: Given the options: ``Activation", ``Inhibition", ``Phosphorylation", ``Dephosphorylation", ``Ubiquitination", ``Deubiquitination", which one is the relation type between WRN and top3 in the text above?\\\\
Answer (true): \textcolor{blue}{Inhibition}
\tcblower
Answer (pred): \textcolor{blue}{Inhibition}
\end{tcolorbox}

For the generation of multiple-choice questions, we identified the six most prevalent categories within the dataset. Subsequently, these categories, comprising \textit{Activation, Inhibition, Phosphorylation, Dephosphorylation, Ubiquitination, and Deubiquitination}, were employed as options for answer choices. The model's performance was evaluated using 500 samples for each class. The results are outlined in Table~\ref{tab:indra-task-1k-eval}, with the corresponding confusion matrices provided in Appendix~\ref{secA7:indra-task-conf-mat}. In general, the larger models demonstrated superior performance compared to the smaller models. Mixtral-8x7B-Instruct (46B) exhibited the highest performance, and SOLAR-Instruct (10.7B) notably achieved the second-best performance, surpassing other larger models. As observed in the confusion matrix, the predictions generated by the Alpaca and Falcon models exhibited significant bias.

\begin{table}[h]
\caption{INDRA DB Gene Regulatory Relation Task - F1 scores with 3,000 samples, consisting of 500 samples from each of the six classes.}\label{tab:indra-task-1k-eval}
\begin{tabular}{@{}p{7cm}cc@{}}
\toprule
 & Micro F1 (\#shot) & Macro F1 (\#shot) \\
\midrule
BioGPT-Large (1.5B) & 0.2267 (0-shot) & 0.1600 (0-shot) \\
BioMedLM (2.7B)     & 0.1443 (0-shot) & 0.1084 (0-shot) \\
Galactica (6.7B)    & 0.5593 (1-shot) & 0.4489 (1-shot) \\
Galactica (30B)     & 0.6560 (1-shot) & 0.5533 (1-shot) \\
Alpaca (7B)	        & 0.1670 (1-shot) & 0.0483 (1-shot) \\
RST (11B)           & 0.4627 (0-shot) & 0.4025 (0-shot) \\
Falcon (7B)         & 0.1707 (1-shot) & 0.0557 (1-shot) \\
Falcon (40B)        & 0.6503 (1-shot) & 0.5494 (1-shot) \\
MPT-Chat (7B)	    & 0.5977 (1-shot) & 0.5105 (1-shot) \\
MPT-Chat (30B)      & 0.6607 (1-shot) & 0.5737 (1-shot) \\
LLaMA2-Chat (7B)	& 0.5767 (1-shot) & 0.5017 (1-shot) \\
LLaMA2-Chat (70B)	& 0.6780 (1-shot) & 0.5906 (1-shot) \\
Mistral-Instruct (7B)	& 0.6380 (1-shot) & 0.5571 (1-shot) \\
Mixtral-8x7B-Instruct (46B)	& \textbf{0.7553} (1-shot) & \textbf{0.6436} (1-shot) \\
SOLAR-Instruct (10.7B)	& 0.7387 (2-shot) & 0.6411 (2-shot) \\
\botrule
\end{tabular}
\footnotetext{Note: \textbf{Bold} indicates the best score.}
\end{table}

As depicted in the confusion matrix, the models exhibited confusion between classes associated with activation and phosphorylation, as well as between inhibition and dephosphorylation. This phenomenon can potentially be elucidated by considering the fundamental role of phosphorylation and dephosphorylation in the regulation of protein function, a process pivotal in determining the activation or inactivation state of a protein. Consequently, the principal outcomes of phosphorylation and dephosphorylation events are often characterized by activation and inhibition, respectively, reflecting the intricate interplay between these biochemical processes. Phosphorylation and dephosphorylation are one mechanism of activation and inhibition. Given the close proximity of these phenomena and the subgroup relations within these classes, the models may encounter challenges in effectively distinguishing between them. This difficulty often results in the models categorizing phosphorylation and dephosphorylation as a broader group, encompassing activation and inhibition, particularly in contexts with limited information. This observation is substantiated by the higher micro F1 scores compared to the macro F1 scores, indicating that the models exhibit more accurate predictions for certain classes than for others.

\section{Conclusion}
This investigation aimed to evaluate the effectiveness of 15 LLMs in the context of various biological tasks, encompassing the identification of PPIs, the recognition of genes associated with human pathways affected by LDR exposure, and the classification of gene regulatory relations. The models were presented with question-answering formatted tasks. In the aggregate, the larger models, namely Mixtral-8x7B-Instruct (46B), SOLAR-Instruct (10.7B), Llama-2-chat (70B), MPT-Chat (30B), and Galactica (30B), demonstrated superior performance, showing promise for specific tasks that involve the extraction of intricate interactions among genes/proteins. While these models contained detailed information for distinct gene/protein groups, they encountered challenges in identifying groups with diverse functions and recognizing gene regulatory relations with high correlations. This suggests a need for additional contextual information and resources to enable accurate responses to such inquiries. Prompt engineering methods, such as RAG (Retrieval Augmented Generation) \cite{lewis2020retrieval}, CoT (Chain-of-Thought) \cite{wei2022chain} or ToT (Tree-of-Thought) \cite{yao2024tree}, ReACT (Reasoning and Acting) \cite{yao2023react}, and DSP (Directional Stimulus Prompting) \cite{li2024guiding}, may facilitate the creation of prompts that incorporate information from external resources. The Parameter Efficient Fine-Tuning (PEFT) of LLMs on downstream tasks also holds the potential to enhance prediction outcomes while minimizing the demand for computational resources and memory. Addressing these aspects remains a focus of our future research endeavors.

\backmatter

\section*{Availability of data and materials}
The code and data are available at: \href{https://github.com/boxorange/BioIE-LLM}{\texttt{https://github.com/boxorange/BioIE-LLM}}.

\section*{Acknowledgements}
This material is based upon works for the Exploration of the Potential for Artificial Intelligence and Machine Learning to Advance Low-Dose Radiation Biology Research (RadBio-AI) program under Award Number DE-SC0012704 and the LUCID: Low-dose Understanding, Cellular Insights, and Molecular Discoveries program under Award Number DE-AC02-06CH1135 supported by the U.S. Department of Energy, Office of Science, Office of Biological and Environmental Research. 

\begin{appendices}

\setcounter{table}{5}
\renewcommand{\thetable}{A\arabic{table}}

\section{Negatome Database description}\label{secA1:db-desc}
The Negatome Database (DB) \citep{blohm2014negatome} stands as a specialized repository dedicated to cataloging Non-Interacting Proteins (NIPs). NIPs are essential for training Protein-Protein Interaction (PPI) prediction algorithms and assessing false positive rates in PPI detection efforts. Negatome DB exhibits lesser bias compared to randomly selected negative data. It encompasses functionally dissimilar interactions, rendering it a robust resource for assessing protein and domain interactions. Data sources were derived through manual curation of literature and analysis of protein complex structures by utilizing an advanced text mining procedure to guide manual annotation. Negatome DB has expanded significantly, growing by over 300\% compared to its initial version, and now contains approximately 6,500 NIPs. Manual verification indicates that nearly half of the text mining results correspond to NIP pairs. The significance of this database lies in its ability to complement positive interaction datasets and aid in refining computational models and assessing PPI predictions.

\section{Tested Prompts}\label{secA2:tested-prompt}

\subsection{STRING DB PPI Task: generative question}

\begin{enumerate}
\setlength{\itemsep}{5pt}
\setlength{\parskip}{0pt}
\item ``Which proteins interact with {x}?"
\item ``Which proteins are related to {x}?"
\item ``Which proteins are bound to {x}?"
\item ``What proteins does {x} bind to?"
\item ``The following proteins are related to {x}"
\item ``The following proteins are bound to the protein {x}"
\item ``The following proteins interact with the protein {x}"
\end{enumerate}

\subsection{STRING/Negatome DB PPI Task: yes/no question }

\begin{enumerate}
\setlength{\itemsep}{5pt}
\setlength{\parskip}{0pt}
\item ``Do {x} and {y} interact with each other?"
\item ``Do the two proteins "{x}" and "{y}" bind each other?"
\item ``Do the two proteins {x} and {y} bind to each other? True or False"
\item ``Do {x} and {y} bind each other? True or False"
\item ``Does {x} bind to {y}? True or False"
\item ``Do {x} and {y} bind to each other? True or False"
\item ``Are {x} and {y} related to each other? yes or no"
\item ``{x} and {y} are related to each other. Is this statement True or False?"
\item ``{x} and {y} are related to each other."
\item ``Given the options: "Related", "Unrelated", which one is the relation type between {x} and {y}?"
\end{enumerate}

\subsection{Pathways affected by LDR Recognition Task}
\begin{enumerate}
\setlength{\itemsep}{5pt}
\setlength{\parskip}{0pt}
\item ``Which genes are associated with "{x}"?"
\item ``Which genes are involved in "{x}"?"
\item ``Which genes are related to {x}?"
\item ``Which genes/proteins are related to {x}?"
\item ``Which molecular objects are associated with "{x}"?"
\end{enumerate}

\subsection{Evaluating Gene Regulatory Relations Task}
\begin{enumerate}
\setlength{\itemsep}{5pt}
\setlength{\parskip}{0pt}
\item ``Which of the following is the relation type between {x} and {y} in the text above?"
\item ``Which of the following is the relation type between "{x}" and "{y}" in the text above?"
\item ``Which of the following is the relation between "{x}" and "{y}" in the text above?"
\item ``Given the options: {z}, which one is the relation type between "{x}" and "{y}" in the text above?"
\item ``Given the options: {z}, which one is the relation type between {x} and {y} in the text above?"
\end{enumerate}

\section{KEGG Pathways affected by LDR exposure Task examples}\label{secA5:kegg-task-pred-ex}

\begin{tcolorbox}[title=A prediction example of Mixtral-8x7B-Instruct-v0.1]
Question: Which genes are involved in "Nicotine addiction”?\\
Answer (pred): \textcolor{blue}{CHRNA4, CHRNB2, CHRNA6, CHRNA7,} \textcolor{red}{CHRNA5, \\CHRNA30, CHRNA26, CHRNA31, CHRNA2,...}
\tcblower
Answer (true): \textcolor{blue}{CHRNA4, CHRNB2, CHRNA6, CHRNA7,} \textcolor{red}{DEE79, DEE27, gluR-B, EIEE79, GABRB2, VGLUT2,...}
\end{tcolorbox}

\begin{tcolorbox}[title=A prediction example of SOLAR-10.7B-Instruct-v1.0]
Question: Which genes are involved in "Phosphonate and phosphinate \\metabolism”?\\
Answer (pred): \textcolor{red}{PNP33, PNP13, PNP14, PNP4, PNP7, PNP40, PNP29,...}
\tcblower
Answer (true): \textcolor{red}{CCTA, SPG82, PCYT1A, CHPT1, CTA, PCYT2, SEPI,...}
\end{tcolorbox}

\begin{tcolorbox}[title=A prediction example of BioGPT-Large]
Question: Which genes are involved in "Mineral absorption”?\\
Answer (pred): \textcolor{blue}{TRPV6, TRPM6, TRPM7,} \textcolor{red}{TRPM5, TRPM4, TRPM1, \\TRPM3}
\tcblower
Answer (true): \textcolor{blue}{TRPV6, TRPM6, TRPM7,} \textcolor{red}{HEPHL1, AHC2, ZFAB, CHAK1, D22S675, STEAP, ATP2B2,...}
\end{tcolorbox}

\begin{tcolorbox}[title=A prediction example of LLaMA-2-7B-Chat]
Question: Which genes are involved in "Antifolate resistance”?\\
Answer (pred): \textcolor{red}{GDF14, GDF17, GDF34, GDF21, GDF36, GDF41, GDF13, GDF38, GDF11, GDF12}
\tcblower
Answer (true): \textcolor{red}{FBP, HGF, MRP1, ABCC1, ZC2HC9, RFC1, NEMO, BCRP1, SHMT, p65}
\end{tcolorbox}

\section{The list of unique KEGG pathways that received full prediction coverage by LLMs}\label{secA6:kegg-task-unique-full-list}
Table~\ref{tab:kegg-task-unique-list} presents a list of unique KEGG pathways for which the models achieved complete prediction coverage.

\begin{table}[h]
\caption{The list of unique pathways that received full prediction coverage by a model}\label{tab:kegg-task-unique-list}
\begin{tabular}{p{4cm} | p{8cm}}
\toprule
Model & Pathways affected by LDR \\
\midrule
BioGPT-Large (1.5B) & `Hippo signaling pathway - multiple species', `Complement and coagulation cascades' \\
\hline
BioMedLM (2.7B)     & `Nitrogen metabolism', `Morphine addiction' \\
\hline
Galactica (6.7B)	& `Valine, leucine and isoleucine degradation', `Mucin type O-glycan biosynthesis' \\
\hline
Galactica (30B)	    & `Arrhythmogenic right ventricular cardiomyopathy', `Fat digestion and absorption', `Glycosaminoglycan biosynthesis - heparan sulfate / heparin', `Intestinal immune network for IgA production', `Tyrosine metabolism', `Glycosaminoglycan biosynthesis - chondroitin sulfate / dermatan sulfate' \\
\hline
Alpaca (7B)	        & `alpha-Linolenic acid metabolism', `Taste transduction', `Linoleic acid metabolism', `Arachidonic acid metabolism' \\
\hline
RST (11B)	        & `MicroRNAs in cancer', `Fluid shear stress and atherosclerosis', `Small cell lung cancer', `Thyroid cancer', `HIF-1 signaling pathway', `Glycerolipid metabolism', `Calcium signaling pathway' \\
\hline
Falcon (7B)	        & `Metabolism of xenobiotics by cytochrome P450', `RNA polymerase' \\
\hline
Falcon (40B)        & `Drug metabolism - cytochrome P450', `GABAergic synapse', `Proteasome' \\
\hline
MPT-Chat (7B)	    & `Neuroactive ligand-receptor interaction' \\
\hline
MPT-Chat (30B)	    & `Necroptosis', `Mismatch repair', `Non-homologous end-joining' \\
\hline
LLaMA2-Chat (7B)	& `Pentose and glucuronate interconversions', `Apoptosis - multiple species', `NF-kappa B signaling pathway' \\
\hline
LLaMA2-Chat (70B)	& `Pancreatic secretion' \\
\hline
Mistral-Instruct (7B)	& `Basal cell carcinoma', `ECM-receptor interaction', `Cholesterol metabolism', `Viral protein interaction with cytokine and cytokine receptor', `Chemical carcinogenesis', `Dilated cardiomyopathy', `Steroid hormone biosynthesis', `Spinocerebellar ataxia'\\
\hline
Mixtral-8x7B-Instruct (46B)	& `One carbon pool by folate', `Maturity onset diabetes of the young', `Glycosaminoglycan degradation', `Phototransduction', `Linoleic acid metabolism', `Primary bile acid biosynthesis', `Arginine biosynthesis' \\
\hline
SOLAR-Instruct (10.7B)	& `Ferroptosis', `Systemic lupus erythematosus', `Ferroptosis' \\
\botrule
\end{tabular}
\end{table}

\section{STRING/Negatome DB PPI Task (yes/no question) confusion matrices}\label{secA4:ppi-confusion-matrix}

\begin{figure}[b]
  \includegraphics[width=\columnwidth]{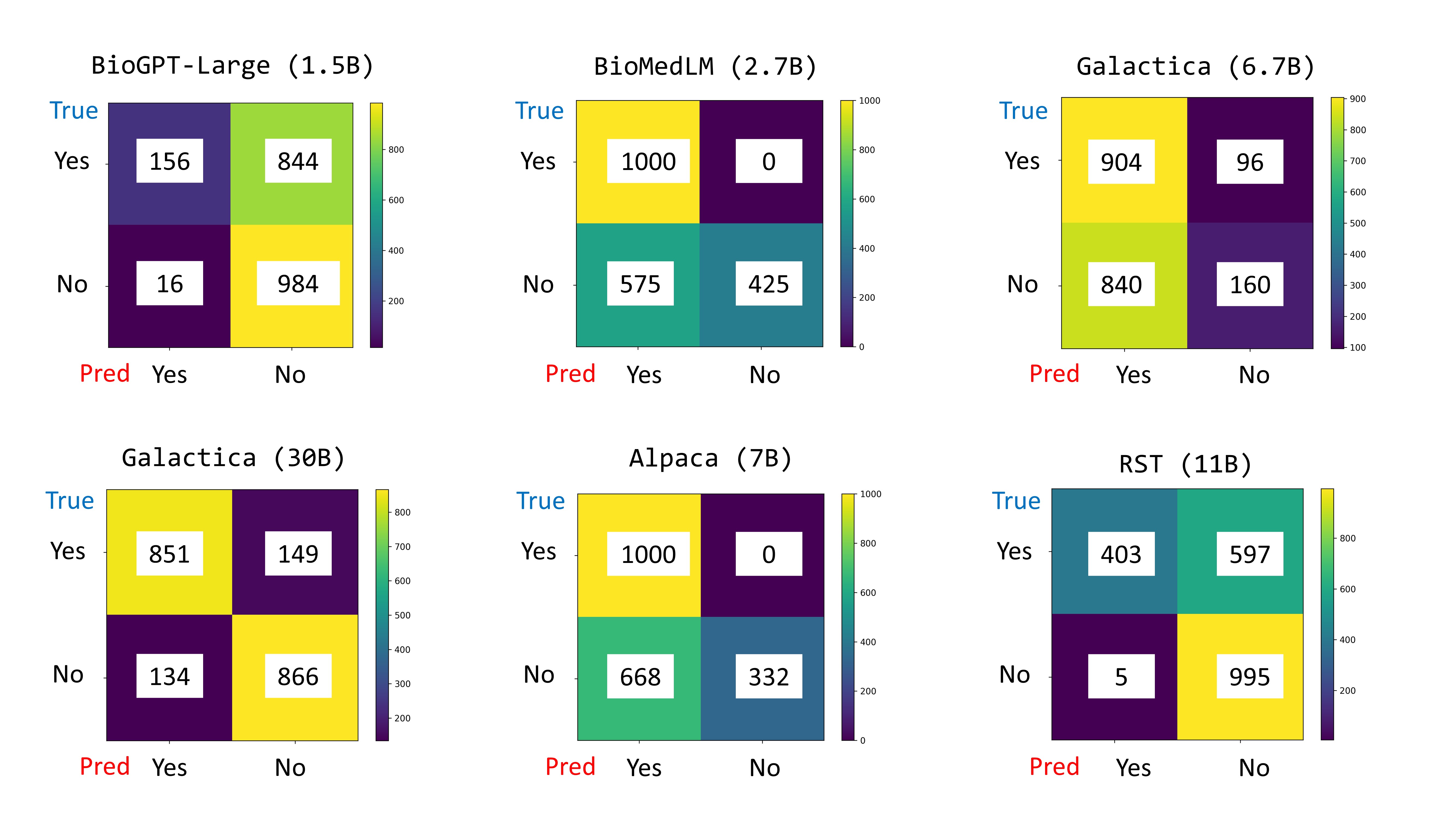}%
  \centering
  \\

  \includegraphics[width=\columnwidth]{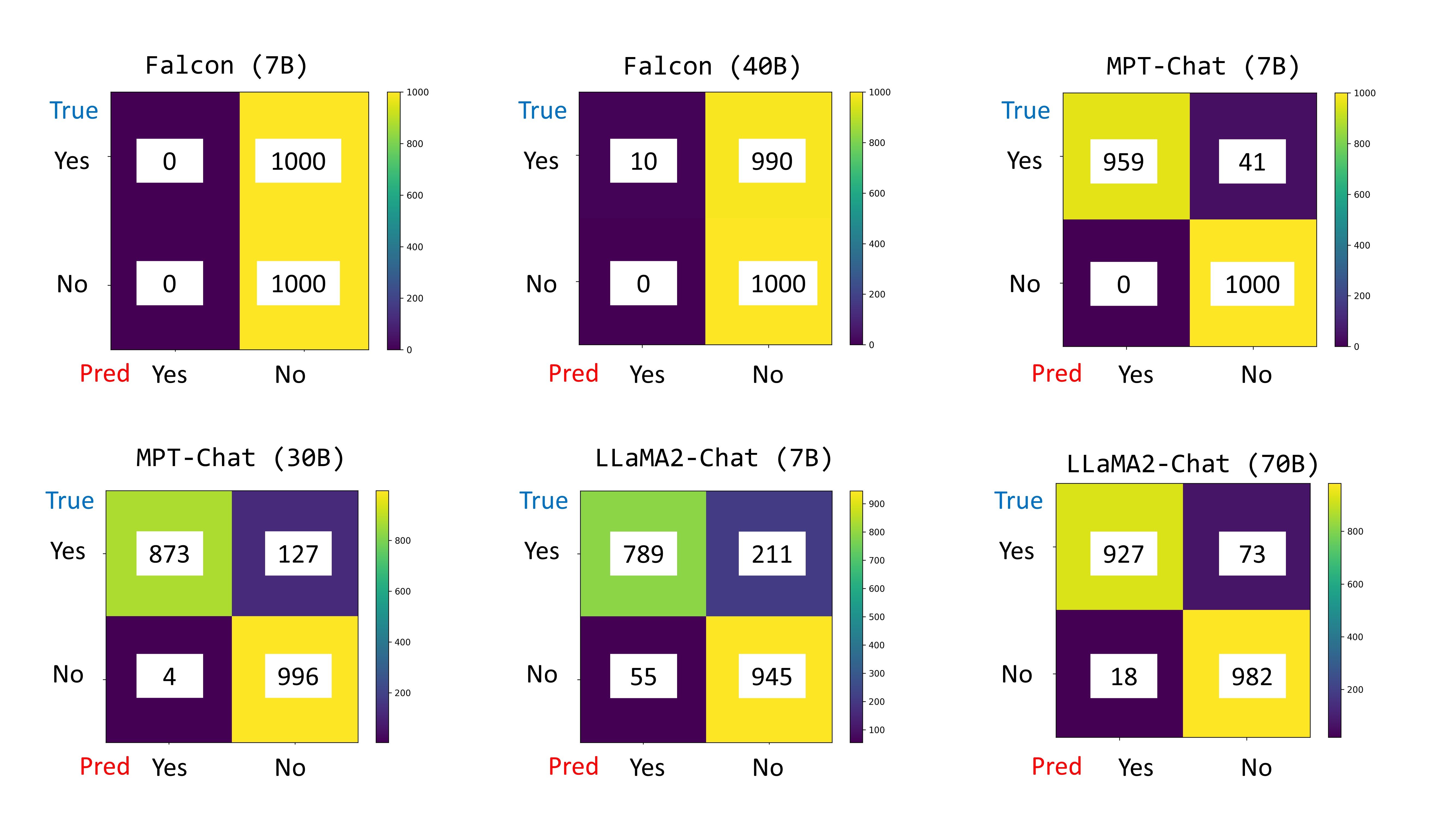}%
  \centering
  \\

  \includegraphics[width=\columnwidth]{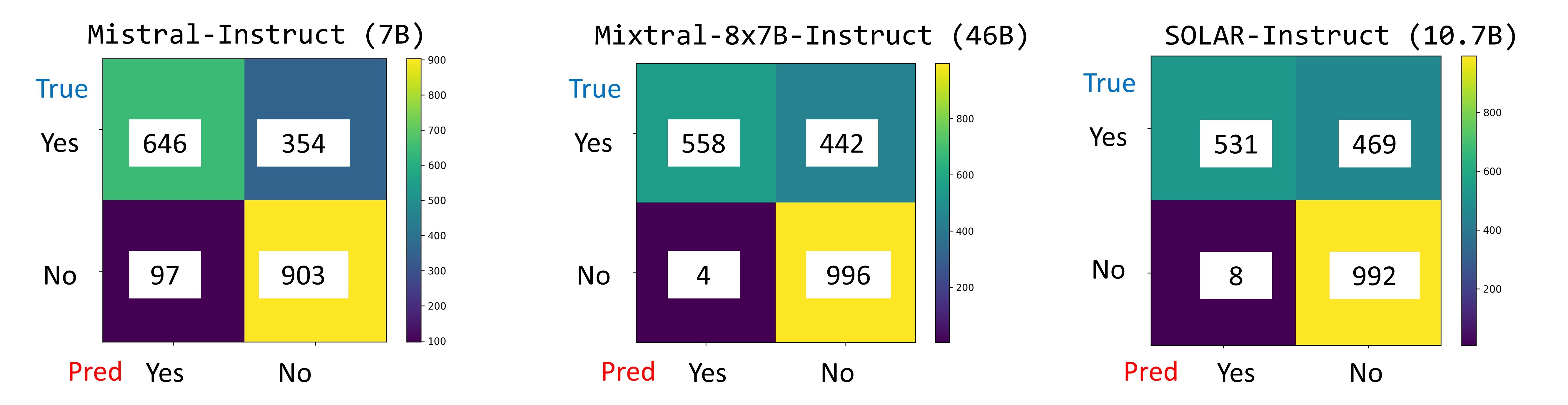}%
  \centering
  \caption{Confusion matrices for STRING/Negatome DB PPI Task (yes/no question).}
  \label{secA4:fig:ppi-confusion-matrix}
\end{figure}

The confusion matrix for the STRING/Negatome DB PPI Task (yes/no question) is presented in Figure~\ref{secA4:fig:ppi-confusion-matrix}.

\section{INDRA DB Gene Regulatory Relation Task (multiple-choice question) confusion matrices}\label{secA7:indra-task-conf-mat}

\begin{figure}
  \includegraphics[width=\columnwidth]{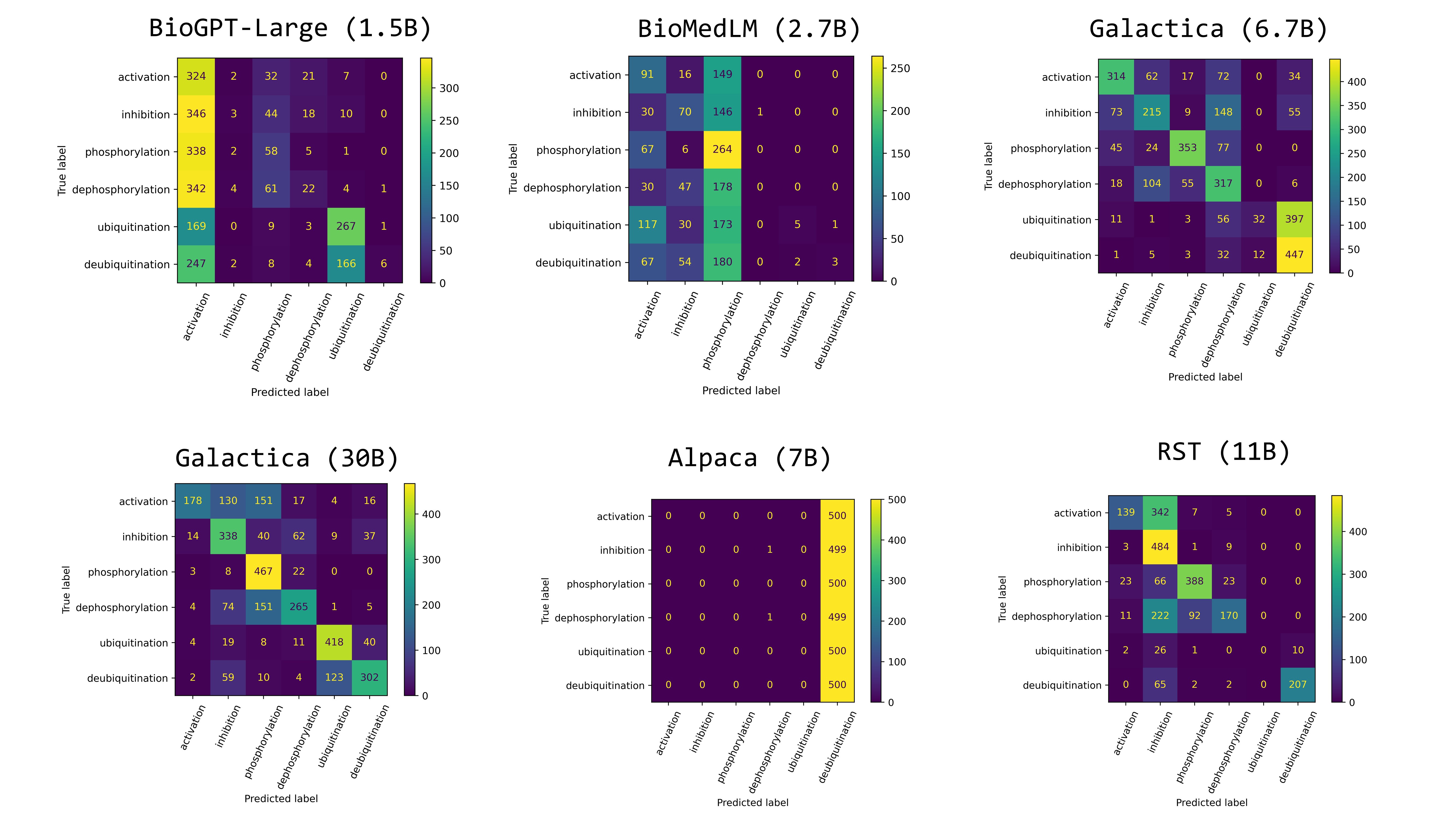}%
  \centering
  \\
  \hfill \break
  \includegraphics[width=\columnwidth]{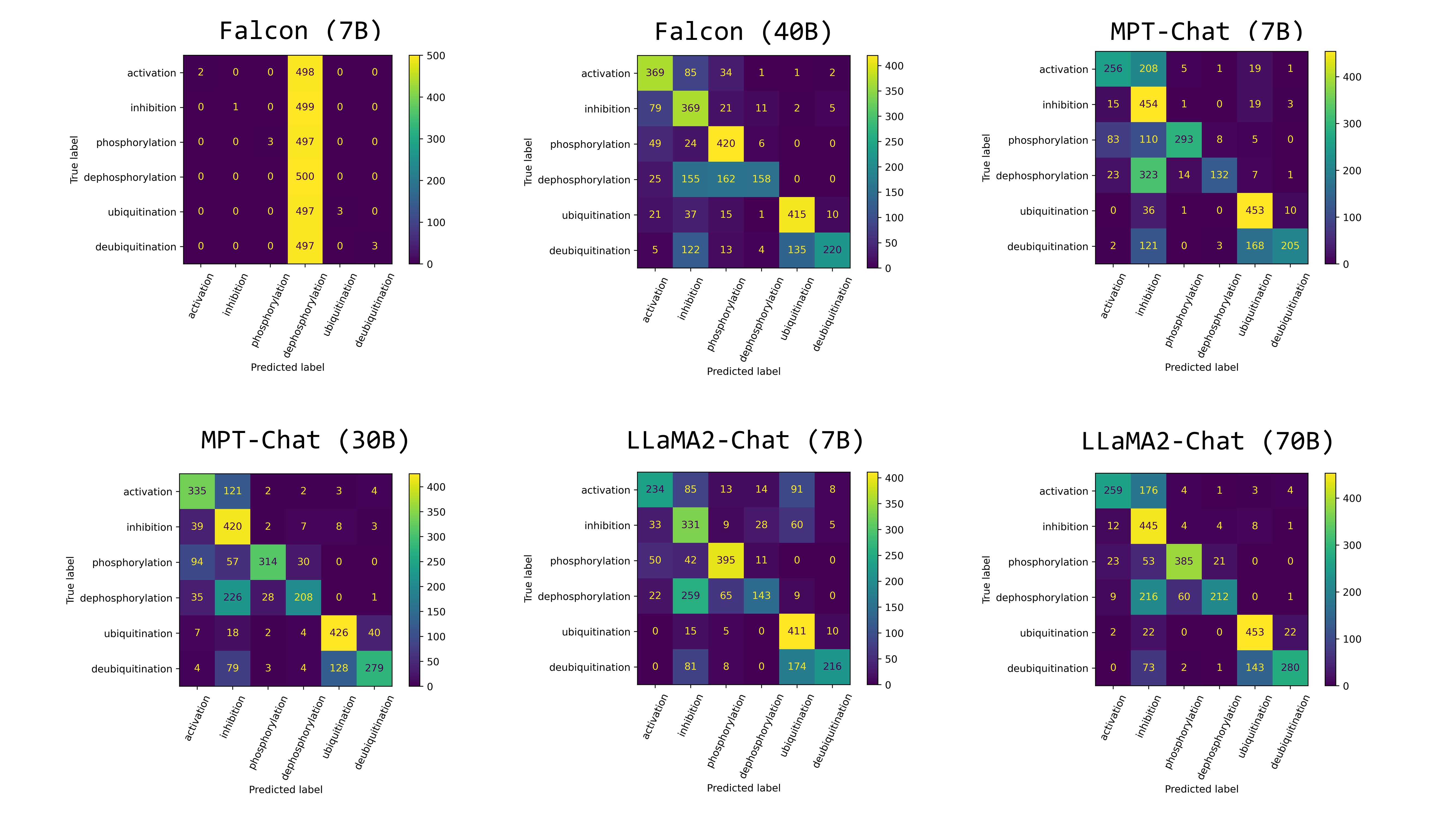}%
  \centering
  \\
  \hfill \break
  \includegraphics[width=\columnwidth]{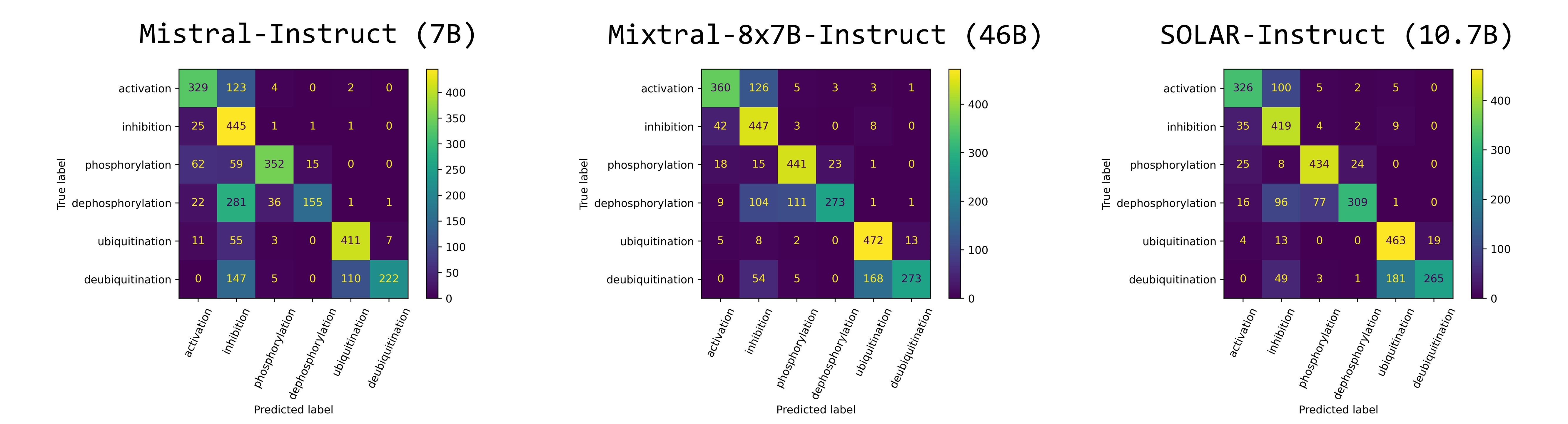}%
  \centering
  \caption{Confusion matrices for INDRA DB Gene Regulatory Relation Task (multiple-choice question).}
  \label{secA7:fig:indra-task-conf-mat}
\end{figure}

The confusion matrix for INDRA DB Gene Regulatory Relation Task (multiple-choice question) is presented in Figure~\ref{secA7:fig:indra-task-conf-mat}.

\section{Task execution duration}\label{secA3:task-exec-time}
The model's task execution time is presented in Table~\ref{secA3:tab:task-exec-time}.

\begin{table}[h]
\caption{Model task execution duration. The experiments were conducted using 4xA100 80GB GPUs. The asterisk (*) next to the model name denotes that model inference was executed across multiple GPUs, with the batch size multiplied by the number of processors.}\label{secA3:tab:task-exec-time}
\begin{tabular}{@{}p{3.5cm}lccc@{}}
\toprule
 & task & \#shot & batch size & time (hh:mm:ss) \\
\midrule
\multirow{4}{*}{BioGPT-Large (1.5B)$\ast$} & STRING PPI & 5 & 32x4=128 & 0:03:09 \\
& STRING\&Negatome PPI & 1 & 64x4=256 & 0:00:37 \\
& KEGG Pathway & 3 & 32x4=128 & 0:00:31 \\
& INDRA Gene Regulatory & 0 & 32x4=128 & 0:00:52 \\
\midrule
\multirow{4}{*}{BioMedLM (2.7B)$\ast$} & STRING PPI & 5 & 32x4=128 & 0:03:04 \\
& STRING\&Negatome PPI & 2 & 64x4=256 & 0:00:52 \\
& KEGG Pathway & 2 & 32x4=128 & 0:00:30 \\
& INDRA Gene Regulatory & 0 & 32x4=128 & 0:01:11 \\
\midrule
\multirow{4}{*}{Galactica (6.7B)$\ast$} & STRING PPI & 5 & 16x4=64 & 0:06:28 \\
& STRING\&Negatome PPI & 1 & 64x4=256 & 0:01:17 \\
& KEGG Pathway & 5 & 4x4=16 & 0:02:15 \\
& INDRA Gene Regulatory & 1 & 8x4=32 & 0:09:06 \\
\midrule
\multirow{4}{*}{Galactica (30B)} & STRING PPI & 5 & 8x1=8 & 3:43:04 \\
& STRING\&Negatome PPI & 5 & 32x1=32 & 0:32:17 \\
& KEGG Pathway & 5 & 4x1=4 & 0:58:45 \\
& INDRA Gene Regulatory & 1 & 4x1=4 & 2:29:51 \\
\midrule
\multirow{4}{*}{Alpaca (7B)$\ast$} & STRING PPI & 5 & 8x4=32 & 0:13:20 \\
& STRING\&Negatome PPI & 5 & 32x4=128 & 0:02:41 \\
& KEGG Pathway & 3 & 8x4=32 & 0:02:18 \\
& INDRA Gene Regulatory & 1 & 4x4=16 & 0:10:10 \\
\midrule
\multirow{4}{*}{RST (11B)} & STRING PPI & 5 & 16x1=16 & 1:49:50 \\
& STRING\&Negatome PPI & 0 & 64x1=64 & 0:01:15 \\
& KEGG Pathway & 2 & 8x1=8 & 0:32:54 \\
& INDRA Gene Regulatory & 0 & 8x1=8 & 0:11:22 \\
\midrule
\multirow{4}{*}{Falcon (7B)$\ast$} & STRING PPI & 5 & 32x4=128 & 0:02:38 \\
& STRING\&Negatome PPI & 1 & 64x4=256 & 0:00:47 \\
& KEGG Pathway & 3 & 8x4=32 & 0:01:10 \\
& INDRA Gene Regulatory & 1 & 8x4=32 & 0:01:51 \\
\midrule
\multirow{4}{*}{Falcon (40B)} & STRING PPI & 5 & 16x1=16 & 1:56:01 \\
& STRING\&Negatome PPI & 1 & 64x1=64 & 0:01:24 \\
& KEGG Pathway & 3 & 8x1=8 & 0:25:14 \\
& INDRA Gene Regulatory & 1 & 4x1=4 & 0:34:19 \\
\midrule
\multirow{4}{*}{MPT-Chat (7B)$\ast$} & STRING PPI & 5 & 16x4=64 & 0:03:44 \\
& STRING\&Negatome PPI & 5 & 32x4=128 & 0:00:51 \\
& KEGG Pathway & 5 & 8x4=32 & 0:01:25 \\
& INDRA Gene Regulatory & 1 & 8x4=32 & 0:01:43 \\
\midrule
\multirow{4}{*}{MPT-Chat (30B)} & STRING PPI & 5 & 16x1=16 & 1:29:55 \\
& STRING\&Negatome PPI & 5 & 32x1=32 & 0:03:32 \\
& KEGG Pathway & 5 & 8x1=8 & 0:27:39 \\
& INDRA Gene Regulatory & 1 & 8x1=8 & 0:21:16 \\
\midrule
\multirow{4}{*}{LLaMA2-Chat (7B)$\ast$} & STRING PPI & 5 & 8x4=32 & 0:12:01 \\
& STRING\&Negatome PPI & 5 & 32x4=128 & 0:03:09 \\
& KEGG Pathway & 5 & 4x4=16 & 0:03:22 \\
& INDRA Gene Regulatory & 1 & 4x4=16 & 0:10:12 \\
\midrule
\multirow{4}{*}{LLaMA2-Chat (70B)} & STRING PPI & 5 & 16x1=16 & 2:55:20 \\
& STRING\&Negatome PPI & 5 & 32x1=32 & 1:17:44 \\
& KEGG Pathway & 5 & 2x1=2 & 1:26:32 \\
& INDRA Gene Regulatory & 1 & 4x1=4 & 5:51:52 \\
\midrule
\multirow{4}{*}{Mistral-Instruct (7B)$\ast$} & STRING PPI & 5 & 8x4=32 & 0:12:23 \\
& STRING\&Negatome PPI & 5 & 32x4=128 & 0:02:51 \\
& KEGG Pathway & 2 & 4x4=16 & 0:02:32 \\
& INDRA Gene Regulatory & 1 & 16x4=64 & 0:10:41 \\
\midrule
\multirow{4}{*}{Mixtral-8x7B-Instruct (46B)} & STRING PPI & 5 & 32x1=32 & 1:02:02 \\
& STRING\&Negatome PPI & 5 & 64x1=64 & 0:15:26 \\
& KEGG Pathway & 2 & 4x1=4 & 0:27:36 \\
& INDRA Gene Regulatory & 1 & 16x1=16 & 1:09:58 \\
\midrule
\multirow{4}{*}{SOLAR-Instruct (10.7B)} & STRING PPI & 5 & 16x1=16 & 4:31:35 \\
& STRING\&Negatome PPI & 3 & 16x1=16 & 0:03:03 \\
& KEGG Pathway & 2 & 8x1=8 & 0:45:55 \\
& INDRA Gene Regulatory & 2 & 16x1=16 & 0:57:46 \\
\botrule
\end{tabular}
\end{table}




\end{appendices}

\clearpage


\bibliography{sn-bibliography}

\end{document}